\documentclass[a4paper,fleqn]{template_cas/cas-dc}
\usepackage[numbers]{natbib}

\usepackage{url}
\usepackage{amsmath}
\usepackage{amsthm}
\usepackage{amssymb}
\usepackage{enumitem}
\usepackage{booktabs}
\usepackage{multirow}
\usepackage{makecell}
\usepackage{xcolor}

\graphicspath{{./}{template_cas/full_extracted_package/els-cas-templates/}}

\hypersetup{
    colorlinks=true,
    linkcolor=red,
    citecolor=blue,
    urlcolor=black,
    filecolor=red
}
\usepackage[capitalize]{cleveref}

\crefname{figure}{Fig.}{Figs.}
\crefname{section}{Section}{Sections}
\crefname{table}{Table}{Tables}
\crefname{algorithm}{Algorithm}{Algorithms}
\crefname{equation}{Eq.}{Eqs.}
\newtheorem{proposition}{\bf Proposition}

\ExplSyntaxOn
\cs_set:Npn \__first_footerline:
{
  \group_begin:
  \small
  \sffamily
  \ifnum\theblind>0\relax
  \else
    \__short_authors: :~
  \fi
  { \rmfamily \itshape Preprint~submitted~to~Pattern~Recognition~Letters }
  \group_end:
}
\ExplSyntaxOff

\begin{document}
\let\WriteBookmarks\relax
\setcounter{topnumber}{2}
\setcounter{totalnumber}{3}
\setlength{\textfloatsep}{8pt plus 2pt minus 2pt}
\setlength{\floatsep}{6pt plus 2pt minus 2pt}
\setlength{\intextsep}{6pt plus 2pt minus 2pt}
\setlength{\dbltextfloatsep}{8pt plus 2pt minus 2pt}
\setlength{\dblfloatsep}{6pt plus 2pt minus 2pt}

\shorttitle{Graph-Indexed Trajectory Patterns for SOTA}
\shortauthors{Wang et~al.}

\title [mode = title]{Learning Graph-Indexed Trajectory Patterns for Stochastic On-Time Arrival Routing}

\author[1]{Yuanhang Wang}
\fnmark[1]
\ead{rayflows42@gmail.com}

\author[1]{Xing Wei}
\fnmark[1]
\ead{cysgynn@gmail.com}

\author[1]{Duoxiang Zhao}
\ead{zhaoduoxiang@stu.scu.edu.cn}

\author[1]{Zezhou Zhang}
\ead{2023141520127@stu.scu.edu.cn}

\author[2,3]{Hao Qin}
\ead{hao.qin@scu.edu.cn}

\author[1]{Yuqi Ouyang}
\cormark[1]
\ead{yuqi.ouyang@scu.edu.cn}

\affiliation[1]{organization={College of Computer Science, Sichuan University},
  city={Chengdu}, postcode={610065}, country={China}}
\affiliation[2]{organization={School of Electrical and Electronic Engineering, University College Dublin},
  city={Dublin}, postcode={D04 V1W8}, country={Ireland}}
\affiliation[3]{organization={College of Electronics and Information Engineering, Sichuan University},
  city={Chengdu}, postcode={610065}, country={China}}

\cortext[1]{Corresponding author.}
\fntext[1]{Yuanhang Wang and Xing Wei contributed equally to this work.}

\begin{abstract}
Correlated link travel times create decision-relevant patterns in partial route histories. In stochastic on-time arrival (SOTA) routing, each route prefix forms a variable-length, graph-indexed sequence in which traversed-edge identities, realized travel times, and route order jointly indicate the reliability of downstream actions. We present GPG-HT, a history-conditioned Transformer policy that learns a trajectory representation from this structured sequence together with the current node, destination, and remaining budget. Edge-time cross-attention and sequence encoding capture dependencies within the observed history, while decoder cross-attention maps the resulting trajectory memory and decision context to an online distribution over feasible outgoing edges. A history-conditioned generalized policy-gradient objective trains the representation from terminal on-time outcomes. Experiments on the Sioux Falls and Anaheim road-network topologies with simulated correlated link times show that GPG-HT achieves higher mean on-time arrival probabilities than representative optimization and reinforcement-learning baselines. Paired common-pool evaluation confirms statistically significant gains in all six network-budget settings, reaching 2.82--3.27 percentage points on Sioux Falls and 0.36--1.04 percentage points on Anaheim. Correlated, independent, shuffled-history, no-history, and architecture controls further demonstrate that GPG-HT learns decision-relevant structure from graph-indexed route prefixes.
\end{abstract}


\maketitle

\section{Introduction}
Correlated stochastic networks yield informative trajectory patterns: early observations update the reliability of downstream actions. In road networks, a realized link travel time conveys evidence about untraversed links through statistical dependence. Each observed route prefix can thus be represented as a variable-length, graph-indexed sequence of edge identities, realized travel times, and route positions.

Online routing therefore calls for a representation that relates this graph-indexed sequence to the current location, destination, and remaining time budget. The Stochastic On-Time Arrival (SOTA) problem \cite{Fan2005} provides this decision setting: each link follows a random travel-time distribution, and the policy maximizes the probability of reaching the destination within a prescribed budget. This deadline-reliability objective complements classical shortest-path criteria based on deterministic or expected travel times \cite{Dijk,AStar} by modeling action selection under travel-time uncertainty.

Research on SOTA spans path-based, adaptive-policy, and learning formulations. Path-based methods pre-compute fixed routes, while policy-based methods select actions from the current routing state. Mathematical programming and dynamic programming provide optimization and Bellman formulations \cite{Cao2016,Yang2017,NieWu2009Reliable,Fan2005,Niknami2016,Shen2017,Prakash2020}. History-adaptive routing conditions policies on realized route information~\cite{Pretolani2009HistoryAdaptive}, and GP4 updates Gaussian travel-time beliefs from observations before selecting the next link~\cite{GaoGuoSheng2021GP4}. Reinforcement-learning methods cast SOTA as a Markov decision process \cite{Cao2020,Guo2022,GuoHeShengEtAl2024SEGAC}, with SEGAC offering a sample-efficient generalized actor-critic solution. These studies establish effective adaptive-routing mechanisms using compact state representations. In parallel, sequence models in tracking and navigation demonstrate the value of learned historical representations~\cite{Zhou2022SpaceTracking,Zhou2023RecurrentTracking,Wang2024UAVNavigation}. Together, these developments motivate a representation that treats the observed edge-time prefix as a structured trajectory pattern while preserving graph position, route order, and the evolving deadline context.

We propose Generalized Policy Gradient with History-aware Transformer (GPG-HT), a graph-indexed trajectory-learning framework for online SOTA routing. GPG-HT encodes the observed edge-time route prefix as trajectory memory and combines it with the current decision context through encoder-decoder attention to predict a probability distribution over feasible next edges. A history-conditioned GPG objective uses terminal on-time outcomes to train the representation over variable-length route histories. This formulation unifies graph-indexed sequential pattern learning and deadline-reliability optimization within a single policy. Our main contributions are:

\begin{itemize}
\item A graph-indexed trajectory representation that organizes traversed-edge identities, realized travel times, and route order as variable-length trajectory memory, with the current node, destination, and remaining budget as decision context.
\item A history-aware Transformer policy with edge-time cross-attention and sequence encoding, trained from terminal on-time outcomes by a history-conditioned GPG objective.
\item Controlled evaluations across two road-network scales include correlated and independent links, shuffled histories, no-history controls, common-pool comparisons, and architecture ablations. The results demonstrate statistically significant SOTA gains and the decision relevance of route-prefix patterns.
\end{itemize}

\begin{figure*}[pos=t]
    \centering
    \includegraphics[width=1\textwidth]{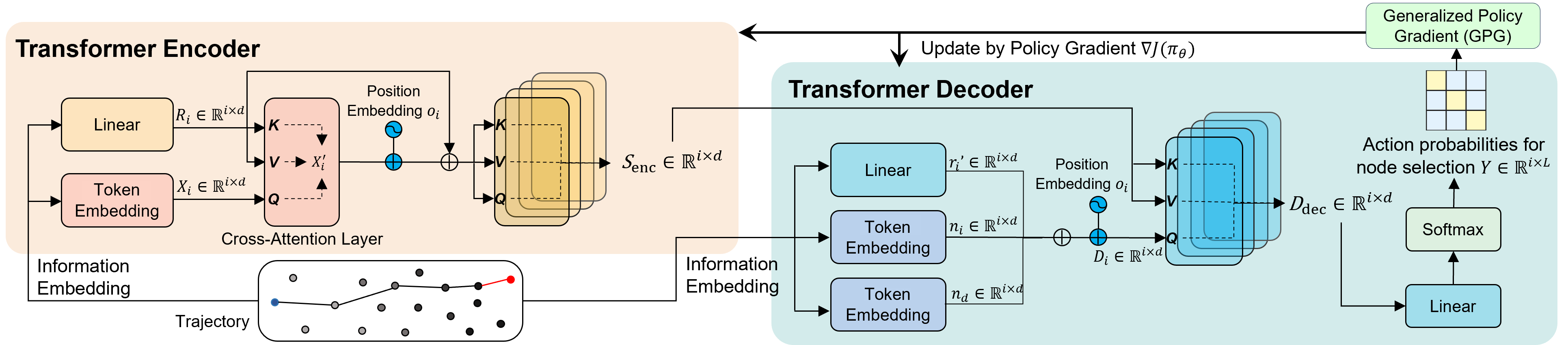}
    \caption{Architecture of GPG-HT. Edge embeddings query realized-time embeddings, while the direct $R_i$ residual preserves the time observation at each route position. Decoder cross-attention combines trajectory memory with decision context to score feasible outgoing edges, indexed in the diagram by their endpoint nodes.}
    \label{fig:model_architecture}
\end{figure*}

\section{History-Conditioned Trajectory Learning}
\Cref{fig:model_architecture} summarizes GPG-HT. The observed route prefix is encoded as trajectory memory, while the current routing constraints form the decision context. Encoder-decoder attention integrates the two representations and predicts a distribution over feasible next edges. A history-conditioned GPG objective trains the policy from terminal on-time outcomes collected through Monte Carlo rollouts.

\subsection{History-Aware Trajectory Representation}
GPG-HT represents variable-length route prefixes with a Transformer encoder-decoder~\cite{Attention}, drawing on sequence-based decision modeling~\cite{Chen2021DecisionTransformer}. Training uses simulated online rollouts with terminal on-time feedback. At decision step $i$, traversed edges $X_i$, realized edge times $R_i$, and route-position encodings $O_i$ form the prefix input, while the current node $n_i$, destination $n_d$, remaining budget $r_i$, and decision-step encoding $o_i$ form the decision context. After projection, $X_i,R_i,O_i\in\mathbb{R}^{m_i\times d}$ describe the $m_i$ observed route positions.

The encoder first applies cross-attention. Edge embeddings query realized-time embeddings to obtain a temporal summary for each observed route position:

\begin{equation}
    X'_i = \operatorname{softmax}\!\left(\frac{X_iR_i^\top}{\sqrt d}\right)R_i.
\label{eq:history_fusion}
\end{equation}
The attention output is combined with route-position encoding $O_i$, while $R_i$ enters every encoder block through a positionwise residual. Cross-attention captures dependencies between each traversed edge and the complete observed time sequence; the residual preserves the realized time aligned with each route position.

The resulting trajectory memory is

\begin{equation}
    S_{\text{enc}}=\operatorname{TransformerBlocks}\!
    \left(X'_i+O_i;\,\mathrm{IR}=R_i\right),
\label{eq:encoder_memory}
\end{equation}
where $\mathrm{IR}$ denotes the additive input residual within the Transformer blocks. Self-attention integrates edge-time interactions across the ordered route prefix to produce $S_{\text{enc}}$, while padding masks accommodate variable-length histories.

The decoder query combines the current node, destination, remaining budget, and decision-step position:

\begin{equation}
    D_i = E_c(n_i)+E_d(n_d)+E_r(r_i)+o_i,
\label{eq:decoder_query}
\end{equation}

Decoder cross-attention then uses $D_i$ as the query and $S_{\text{enc}}$ as the key/value trajectory context:

\begin{equation}
    D_{\text{dec}} = \text{TransformerDecoder}\!\left(D_i,S_{\text{enc}}\right),
\label{eq:decoder_attention}
\end{equation}

\begin{equation}
    Y_i = \operatorname{softmax}\!\left(\mathcal{M}_{n_i}\!\left(\operatorname{FC}(D_{\text{dec}})\right)\right),
\label{eq:output_layer}
\end{equation}
where $\mathcal{M}_{n_i}$ masks edges that do not leave $n_i$, and $\operatorname{FC}$ denotes the output layer. The output $Y_i$ is a probability distribution over feasible outgoing edges, indexed in \cref{fig:model_architecture} by endpoint node. After edge $a_i$ is sampled and travel time $t_{i+1}$ is observed, they are appended to $X_i$ and $R_i$, respectively, and the budget is updated as $r_{i+1}=r_i-t_{i+1}$. Iteration continues until arrival or a terminal failure state, yielding a Monte Carlo trajectory for the history-conditioned GPG estimator.

\begin{table}[pos=t]
\centering
\caption{Classical SOTA and GPG-HT formulation mapping.}
\label{tab:rl_mapping}
\small
\setlength{\tabcolsep}{3pt}
\resizebox{\columnwidth}{!}{%
\begin{tabular}{@{}lll@{}}
\toprule
Element & Classical Markovian SOTA & GPG-HT \\
\midrule
State & Current node and budget & History-aware trajectory state $(h_i,c_i)$ \\
Action & Feasible outgoing edge & Masked next-edge distribution $Y_i$ \\
Reward & On-time indicator & $\mathbf{1}\{G(\xi)\le T\}$ \\
Policy & DP/RL rule & Transformer policy $\pi_\theta$ \\
\bottomrule
\end{tabular}}
\end{table}

\subsection{History-Conditioned Generalized Policy Gradient}
The history-aware Transformer parameterizes the policy $\pi_\theta$ over graph-indexed trajectory states. At decision step $i$, the indexed edge-time history $h_i$ and decision context $c_i$ define the policy input:

\begin{equation}
 h_i = (X_i,R_i), \qquad c_i = (r_i,n_i,n_d,o_i),
\label{eq:key}
\end{equation}

\begin{equation}
\mathbb{P}_\theta(a_i\mid h_i,c_i) = \pi_\theta(a_i\mid h_i,c_i),
\label{eq:AC}
\end{equation}
Here, $a_i$ is the selected outgoing edge, $h_i$ is the indexed edge-time history, and $c_i$ is the decision context. \Cref{tab:rl_mapping} summarizes the correspondence between this representation and classical reinforcement-learning and shortest-path formulations. The $j$th trajectory generated by $\pi_\theta$ is defined as:

\begin{equation}
\begin{aligned}
    \xi_j =& (h_{j,0},c_{j,0},a_{j,0},t_{j,1},h_{j,1},c_{j,1},\dots,\\
    &h_{j,i_j-1},c_{j,i_j-1},a_{j,i_j-1},t_{j,i_j},h_{j,i_j},c_{j,i_j}).
    \label{eq:trajectory}
\end{aligned}
\end{equation}

\begin{equation}
G(\xi_j) = \sum_{k=1}^{i_j} t_{j,k},
\end{equation}
where $i_j$ is the number of decisions and $t_{j,k}$ is the realized travel time after decision $k-1$. The objective maximizes the probability that total travel time does not exceed budget $T$:

\begin{equation}
J(\pi_\theta) = \mathbb{P}[G(\xi_{j}) \le T].
\label{eq:GPG_obj}
\end{equation}
The terminal on-time event provides a trajectory-level reliability signal. Applying the standard likelihood-ratio identity \cite{VanillaPolicyGradient} to the history-conditioned policy yields the following gradient of \cref{eq:GPG_obj}, with each action log-probability evaluated on the graph-indexed route prefix observed at that step:

\begin{equation}
    \nabla_{\theta} J(\pi_\theta) = \mathbb{E}_{\xi_j \sim p_\theta}
    \Bigl[
    \mathbf{1}\{G(\xi_j) \leq T\} \cdot
    \nabla_{\theta} \log p_\theta(\xi_j)
    \Bigr],
\label{eq:GPG}
\end{equation}
The sampled on-time indicator supplies the terminal trajectory label, and Monte Carlo averaging estimates the expectation used to update the Transformer policy. The following proposition gives the corresponding estimator for variable-length route histories and realized travel times.

\begin{proposition}[Monte Carlo estimation of history-conditioned GPG]
    The Monte Carlo approximation of \cref{eq:GPG} with $M$ sampled trajectories $\{\xi_{j}\}_{j=1}^M$ generated by policy $\pi_\theta$ is:
    \small
    \begin{equation}
        \nabla_{\theta} J(\pi_\theta)
        \approx \frac{1}{M} \sum_{j=1}^M
        \mathbf{1}\{G(\xi_j) \leq T\} \cdot
        \sum_{k=0}^{i_j-1} \nabla_{\theta}
        \log \pi_\theta(a_{j,k} | h_{j,k}, c_{j,k}),
    \label{eq:Monte_Carlo_GPG}
    \end{equation}
    \normalsize
where $\displaystyle \mathbf{1}\{\cdot\}$ is the indicator function and $i_j$ is the number of decision steps in the $j_{\text{th}}$ trajectory.
\end{proposition}

\begin{proof}
For compactness, write $s_{j,k}=(h_{j,k},c_{j,k})$ and let
$p^{\mathrm{env}}_{j,k}$ denote the environment transition probability
from $(s_{j,k},a_{j,k})$ to the next sampled travel time and state.
For a trajectory $\xi_j$, its log-probability is
\begin{equation}
\begin{aligned}
\log p_\theta(\xi_j)
&= \sum_{k=0}^{i_j-1}
\Bigl[
\log \pi_\theta(a_{j,k}\mid s_{j,k}) \\
&\qquad\qquad + \log p^{\mathrm{env}}_{j,k}
\Bigr].
\end{aligned}
\end{equation}
Because the environment transition dynamics are independent of $\theta$,
their gradient vanishes. Therefore,
\begin{equation}
\begin{aligned}
\nabla_\theta \log p_\theta(\xi_j)
&= \sum_{k=0}^{i_j-1}
\nabla_\theta \log \pi_\theta(a_{j,k}\mid s_{j,k}).
\end{aligned}
\end{equation}
Substituting this identity into the trajectory-level policy-gradient
objective gives
\begin{equation}
\begin{aligned}
\nabla_\theta J(\pi_\theta)
&= \mathbb{E}_{\xi_j\sim p_\theta}
\Bigl[
\mathbf{1}\{G(\xi_j)\leq T\} \\
&\qquad \times
\sum_{k=0}^{i_j-1}
\nabla_\theta \log \pi_\theta(a_{j,k}\mid s_{j,k})
\Bigr].
\end{aligned}
\end{equation}
Approximating the expectation by the empirical mean over $M$ sampled
trajectories $\{\xi_j\}_{j=1}^M$ yields \cref{eq:Monte_Carlo_GPG}.
This yields the Monte Carlo estimator used by GPG-HT. Because every policy
term is evaluated on $(h_{j,k},c_{j,k})$, each gradient contribution is
conditioned on the observed route prefix, current routing state, and budget
at that decision step.
\end{proof}
During training, each mini-batch aggregates terminal on-time indicators across simulated trajectories. The resulting updates associate edge-time prefix patterns and budget evolution with terminal reliability, while held-out rollout reliability selects the final checkpoint.

\section{Experiments}
Experiments evaluate whether route prefixes contain predictive structure, whether GPG-HT uses it, and whether gains persist across graph scales and held-out realizations. Controlled studies identify the information source, architecture ablations quantify representation components, and paired common-pool comparisons measure routing performance.

\subsection{Data}
We evaluate on two standard US road-network benchmark topologies: the Sioux Falls Network (SFN) \cite{LEBLANC1975309} with 24 nodes and 76 directed links, and the Anaheim Network (AN) \cite{GuoHeShengEtAl2024SEGAC} with 416 nodes and 914 links. The topologies are road-network benchmarks, while link travel times are simulated stochastic variables whose marginal means and variances follow the benchmark settings~\cite{GuoHeShengEtAl2024SEGAC}. SFN uses one fixed full covariance matrix across all data splits. For AN, pairwise coefficients are drawn from $U(-1,1)$, symmetrized, projected onto the positive-semidefinite cone, and diagonally renormalized; the resulting covariance is fixed before sample-pool generation. Each episode draws one joint Gaussian link-time vector, with sampled times floored at 0.1. A traversed edge reveals its realized time and appends it to the route history, so correlated observations can carry information about downstream alternatives. Training, checkpoint selection, and final evaluation use disjoint fixed sample pools.

\begin{figure*}[pos=t]
    \centering
    \includegraphics[width=0.9\textwidth]{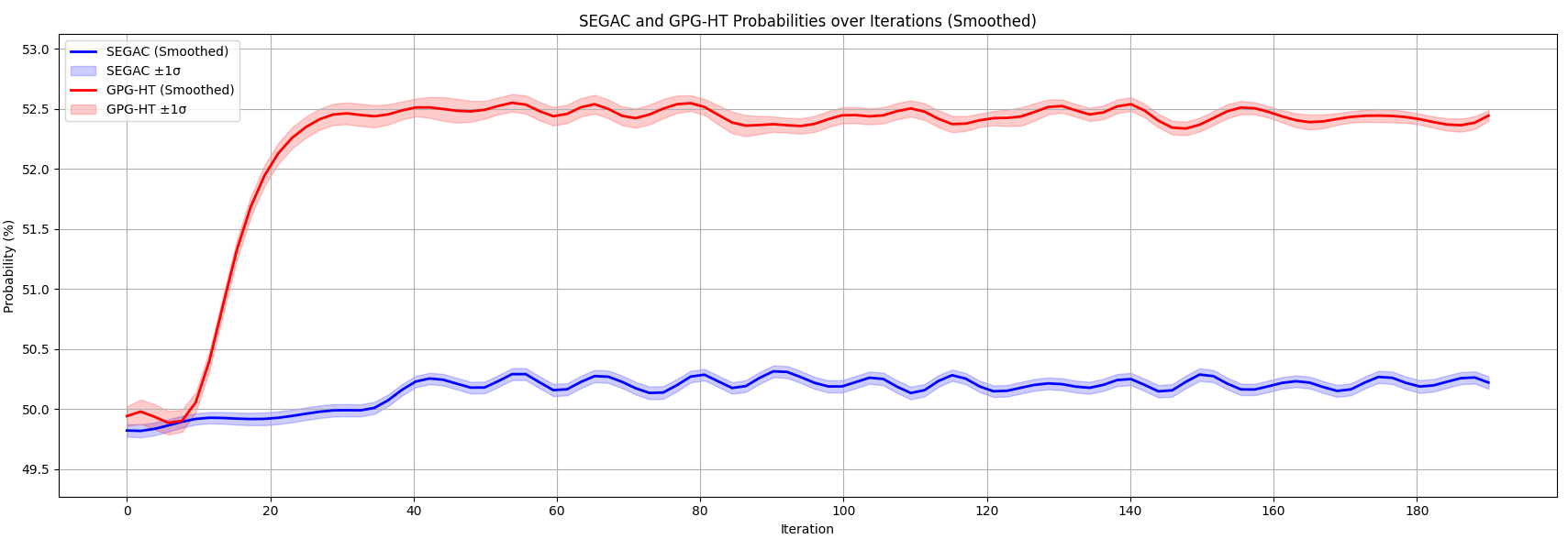}
    \caption{Controlled synthetic validation. GPG-HT conditions on route history and approaches the computed adaptive SOTA upper bound more closely than SEGAC.}
    \label{fig:gpg_dh_learning}
\end{figure*}

\subsection{Implementation Details}
GPG-HT uses 256-dimensional embeddings, four Transformer layers, and eight attention heads. It is optimized with AdamW~\cite{Loshchilov2019AdamW} on fixed stochastic sample pools, and checkpoints are selected by held-out rollout evaluation. The six baselines follow their benchmark formulations and reported optimized settings~\cite{Fan2005,shahabi2015robust,chen2013reliable,GuoHeShengEtAl2024SEGAC}; each solver or training configuration is fixed before final evaluation.

We evaluate five OD pairs under three budgets, $T=0.95, 1.00, 1.05$ times the LET path travel time computed by Dijkstra's algorithm on mean link times. The SFN OD pairs are 2-15, 4-7, 10-13, 13-19, and 17-24; the AN OD pairs are 96-161, 43-161, 376-52, 34-32, and 402-47. We compare with DOT \cite{Prakash2020}, FMA \cite{GuoHeGaoRus2022TimeLimits}, MLR \cite{Yang2017}, PQL \cite{Cao2020}, CTD \cite{Guo2022}, and SEGAC \cite{GuoHeShengEtAl2024SEGAC}. All methods are evaluated using the same OD pairs, budgets, and ten held-out common stochastic-realization pools, each containing 2,000 joint realizations; within each pool, results are averaged equally over the five OD pairs. Non-outgoing edges are masked. Revisited nodes are allowed within a 12-step SFN or 32-step AN cap; budget exhaustion, no feasible outgoing edge, or an unfinished rollout at the cap counts as failure.

The baselines cover DP (DOT/FMA), mixed-integer optimization (MLR), RL (PQL/CTD), and sample-efficient actor-critic SOTA (SEGAC). Together they represent optimization, DP, and RL families. The tables report mean on-time arrival probabilities across the common pools.

\subsection{Evaluation Metrics}
Performance is measured by the SOTA probability $J$, i.e., the fraction of simulated paths arriving within budget:

\begin{equation}
   J = \frac{1}{M} \sum_{j=1}^M \displaystyle \mathbf{1}\{G(\xi_j) \leq T\},
\end{equation}
where $G(\xi_j)$ is the travel time of path $\xi_j$, $T$ is the budget, and $M$ is the number of simulated paths. This metric captures deadline reliability under stochastic travel times. Each common-pool mean is treated as an independent statistical unit. Two-sided paired Student $t$ intervals and tests use the ten poolwise differences ($9$ degrees of freedom), and Holm correction controls familywise error across each reported family of comparisons. These statistics quantify evaluation-pool uncertainty conditional on the selected checkpoints.

\begin{table}[pos=t]
\centering
\caption{Controlled history-dependence validation. Values are mean on-time arrival probabilities over 20 independent stochastic-sampling seeds with 100,000 Monte Carlo samples per seed.}
\label{tab:history_control}
\small
\setlength{\tabcolsep}{3pt}
\begin{tabular}{@{}llcc@{}}
\toprule
Setting & Decision signal & Mean & Gain \\
\midrule
Correlated links & Realized history & 0.5292 & +0.0300 \\
Correlated links & No history & 0.4992 & -- \\
Correlated links & Shuffled history & 0.4994 & +0.0002 \\
Independent links & Realized history & 0.4999 & +0.0003 \\
Independent links & No history & 0.4997 & -- \\
\bottomrule
\end{tabular}
\end{table}

\begin{table}[pos=t]
\centering
\caption{Mean on-time arrival probabilities on the SFN benchmark over ten common held-out evaluation pools. Best results are highlighted in bold.}
\label{tab:algorithm_performance_sfn_mean}
\small
\setlength{\tabcolsep}{4pt}
\resizebox{\columnwidth}{!}{%
\begin{tabular}{@{}cccccccc@{}}
\toprule
T & GPG-HT (Ours) & DOT & FMA & MLR & PQL & CTD & SEGAC \\
\midrule
0.95 & \textbf{0.4309} & 0.3880 & 0.4019 & 0.3934 & 0.3811 & 0.4008 & 0.3928 \\
1.00 & \textbf{0.5266} & 0.4971 & 0.4984 & 0.4955 & 0.4865 & 0.4945 & 0.4972 \\
1.05 & \textbf{0.6703} & 0.6229 & 0.6266 & 0.6256 & 0.6313 & 0.6376 & 0.6376 \\
\bottomrule
\end{tabular}}
\end{table}

\begin{figure}[pos=t]
    \centering
    \includegraphics[width=\columnwidth]{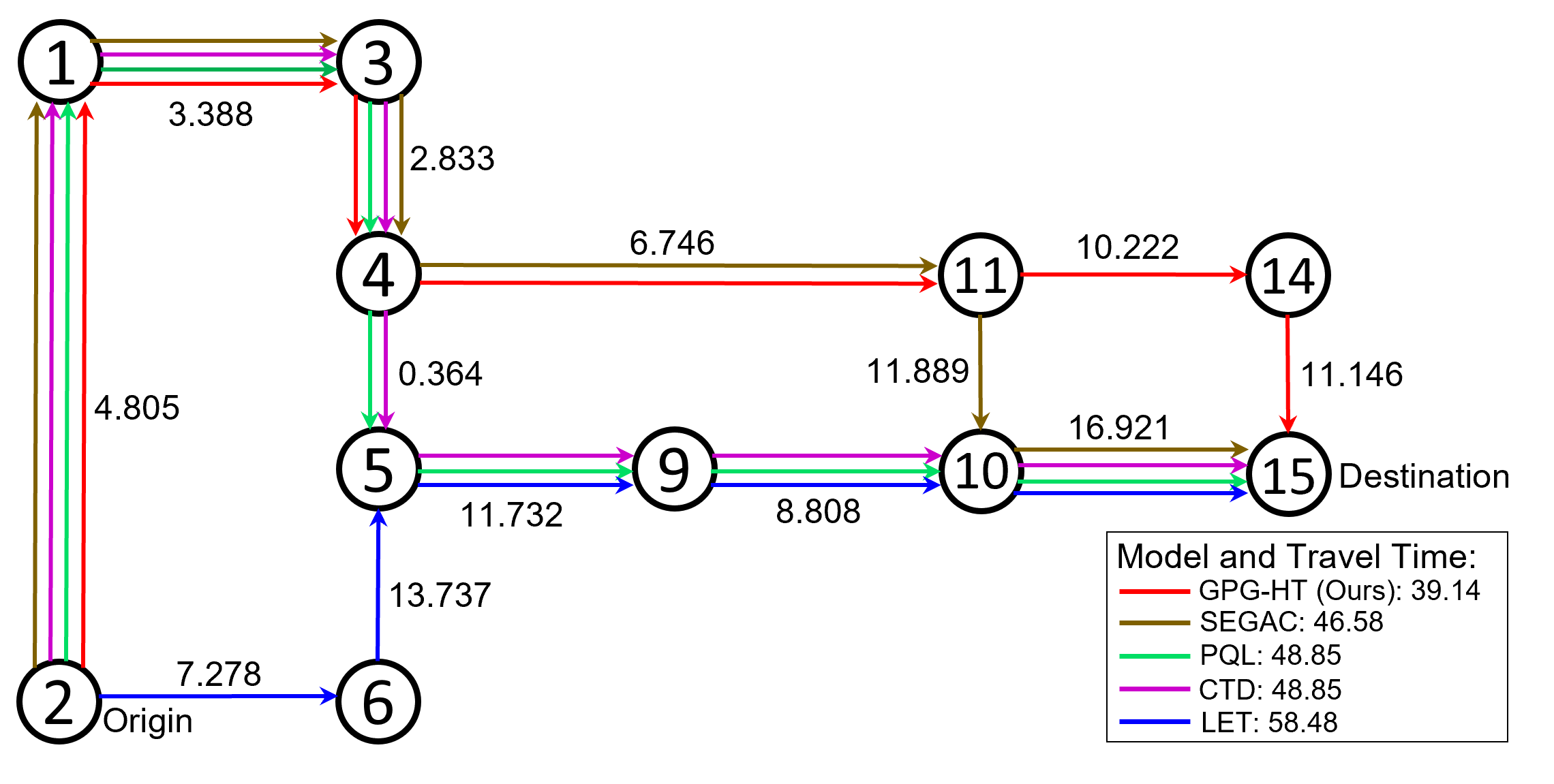}
    \caption{A travel example for OD pair 2-15 on SFN with time budget $T=0.95$. Colored arrows indicate route choices under different methods.}
    \label{fig:Sioux_travel}
\end{figure}

\begin{table}[pos=t]
\centering
\caption{Mean on-time arrival probabilities on the AN benchmark over ten common held-out evaluation pools. Best results are highlighted in bold.}
\label{tab:algorithm_performance_anaheim}
\small
\setlength{\tabcolsep}{4pt}
\resizebox{\columnwidth}{!}{%
\begin{tabular}{@{}cccccccc@{}}
\toprule
T & GPG-HT (Ours) & DOT & FMA & MLR & PQL & CTD & SEGAC \\
\midrule
0.95 & \textbf{0.3021} & 0.2650 & 0.2616 & 0.2632 & 0.2758 & 0.2671 & 0.2985 \\
1.00 & \textbf{0.5439} & 0.4975 & 0.4963 & 0.4948 & 0.5335 & 0.5088 & 0.5246 \\
1.05 & \textbf{0.7956} & 0.7700 & 0.7872 & 0.7752 & 0.7899 & 0.7844 & 0.7899 \\
\bottomrule
\end{tabular}}
\end{table}

\begin{table}[pos=t]
\centering
\caption{Ablated mean on-time arrival probabilities on the SFN benchmark under three time budgets. HA: history-aware; DT: Transformer decision module; GPG: generalized policy gradient.}
\label{tab:AblationCombined}
\small
\setlength{\tabcolsep}{2pt}
\renewcommand{\arraystretch}{0.96}
\resizebox{0.9\columnwidth}{!}{%
\begin{tabular}{@{}ccccccccc@{}}
\toprule
\multirow{2}{*}{HA} & \multirow{2}{*}{DT} & \multirow{2}{*}{GPG} & \multirow{2}{*}{T} & \multicolumn{5}{c}{Origin-Destination Pairs} \\ 
\cmidrule(lr){5-9}
 & & & & 2-15 & 4-7 & 10-13 & 13-19 & 17-24 \\
\midrule
\multirow{3}{*}{$\times$} & \multirow{3}{*}{\checkmark} & \multirow{3}{*}{\checkmark} & 0.95 & 0.4054 & 0.4390 & 0.3984 & 0.4143 & 0.4849 \\
 & & & 1.00 & 0.4989 & 0.5459 & 0.4968 & 0.5079 & 0.5733 \\
 & & & 1.05 & 0.6913 & 0.6500 & 0.6082 & 0.7160 & 0.6663 \\
\midrule
\multirow{3}{*}{\checkmark} & \multirow{3}{*}{$\times$} & \multirow{3}{*}{\checkmark} & 0.95 & 0.4054 & 0.4397 & 0.3984 & 0.4146 & 0.4868 \\
 & & & 1.00 & 0.4990 & 0.5459 & 0.4968 & 0.5064 & 0.5732 \\
 & & & 1.05 & 0.6906 & 0.6500 & 0.6082 & 0.7192 & 0.6675 \\
\midrule
\multirow{3}{*}{\checkmark} & \multirow{3}{*}{\checkmark} & \multirow{3}{*}{$\times$} & 0.95 & 0.4059 & 0.4410 & 0.3984 & 0.4183 & 0.4917 \\
 & & & 1.00 & 0.4978 & 0.5459 & 0.4964 & 0.5171 & 0.5754 \\
 & & & 1.05 & 0.6909 & 0.6511 & 0.6082 & 0.7278 & 0.6735 \\
\midrule
\multirow{3}{*}{\checkmark} & \multirow{3}{*}{\checkmark} & \multirow{3}{*}{\checkmark} & 0.95 & 0.4060 & 0.4411 & 0.3984 & 0.4184 & 0.4906 \\
 & & & 1.00 & 0.4981 & 0.5459 & 0.4963 & 0.5177 & 0.5751 \\
 & & & 1.05 & 0.6912 & 0.6513 & 0.6082 & 0.7274 & 0.6734 \\
\bottomrule
\end{tabular}}
\end{table}

\subsection{Predictive Structure in Route Histories}
Before the benchmark comparison, we use a five-node synthetic network to isolate the information carried by a route prefix. Its two routes, $A\!\to\!B\!\to\!C\!\to\!E$ and $A\!\to\!B\!\to\!D\!\to\!E$, share edge $AB$. The travel times of $AB$, $BC$, and $BD$ are denoted by $(X_1,X_2,X_3)$, while $CE$ and $DE$ each have fixed travel time 1. The random travel times follow a multivariate normal distribution with
\begin{equation}
    \mu=[5,100,100],\quad
    \Sigma =
    \begin{bmatrix}
    1 & 0.5 & 0 \\
    0.5 & 2 & -1 \\
    0 & -1 & 2
    \end{bmatrix}.
\end{equation}
In this construction, the realized time on the shared edge changes the conditional reliability of downstream alternatives, providing a direct test of route-prefix representation learning. The adaptive upper bound conditions on the observed $X_1$ and selects at node $B$ the branch with the higher conditional on-time probability. \Cref{fig:gpg_dh_learning} shows GPG-HT converging to about 52.6\% SOTA probability, approaching the computed upper bound of 53.7\% and exceeding SEGAC at 50.2\%. This ordering demonstrates that the learned route-prefix representation recovers decision-relevant structure from early observations.

A matched control study tests whether the gain follows the observation-position correspondence. On the same two-branch network, the realized-history rule uses the observed shared-edge time to select the downstream alternative, the no-history rule selects a fixed alternative from independent samples, and history shuffling disrupts the correspondence between early observations and later choices. \Cref{tab:history_control} shows that realized history under correlated links yields a 3.0 percentage-point gain with a paired 95\% CI of [0.0295, 0.0305]. History shuffling reduces the gain to 0.0002, while realized history under independent links yields 0.0003 with a 95\% CI of [-0.0001, 0.0006]. Together, the results identify link-time correlation and preserved route position as the sources of predictive value.

\subsection{Performance on the Sioux Falls Network}

Table~\ref{tab:algorithm_performance_sfn_mean} reports mean on-time arrival probabilities on SFN over five OD pairs and three travel-time budgets. GPG-HT ranks first at every budget, with gains of 0.0282--0.0327, or 2.82--3.27 percentage points, over the strongest baseline in each setting. The largest margin occurs at $T=1.05$, where GPG-HT reaches 0.6703 and both CTD and SEGAC reach 0.6376. These results align with the controlled pattern-learning evidence: the route prefix encodes accumulated budget consumption and correlation-dependent signals about downstream links. \Cref{fig:Sioux_travel} illustrates this process for OD pair 2-15 at $T=0.95$. After observing the realized travel times along \textit{2$\to$1$\to$3$\to$4}, GPG-HT selects \textit{4$\to$11} and subsequently follows \textit{11$\to$14$\to$15}, producing the lowest travel time among the methods in the illustrated realization.

\subsection{Performance on the Anaheim Network}
Table~\ref{tab:algorithm_performance_anaheim} reports mean on-time arrival probabilities on AN over five OD pairs and three budgets. GPG-HT ranks first at every budget on this larger road-network topology. SEGAC is the strongest baseline at $T=0.95$, and PQL is strongest at $T=1.00$ and $T=1.05$; the corresponding GPG-HT gains range from 0.00359 to 0.01042. Paired common-pool tests confirm statistical significance in all three settings. Together with the SFN results, the AN evaluation establishes cross-scale consistency for the graph-indexed route-prefix representation.

\begin{table}[pos=t]
\centering
\caption{Paired common-pool comparisons using ten pool-level differences. Panel (a) compares GPG-HT with the strongest baseline in each setting. Panel (b) reports full-model gains over SFN architecture ablations; all six component tests remain significant after Holm correction (maximum adjusted $p=0.0103$).}
\label{tab:paired_stats}
\small
\setlength{\tabcolsep}{2pt}
\textit{(a) Strongest external baseline in each setting.}\par\smallskip
\resizebox{\columnwidth}{!}{%
\begin{tabular}{@{}lccccc@{}}
\toprule
Network & T & Baseline & Gain & Paired 95\% CI & Raw $p$ \\
\midrule
SFN & 0.95 & FMA & 0.0290 & [0.0267, 0.0313] & $3.9\times10^{-10}$ \\
SFN & 1.00 & FMA & 0.0282 & [0.0248, 0.0316] & $1.6\times10^{-8}$ \\
SFN & 1.05 & CTD/SEGAC & 0.0327 & [0.0290, 0.0364] & $8.7\times10^{-9}$ \\
AN & 0.95 & SEGAC & 0.0036 & [0.0012, 0.0060] & 0.0076 \\
AN & 1.00 & PQL & 0.0104 & [0.0065, 0.0143] & $1.9\times10^{-4}$ \\
AN & 1.05 & PQL & 0.0057 & [0.0024, 0.0090] & 0.0034 \\
\bottomrule
\end{tabular}}
\par\smallskip
\textit{(b) SFN architecture ablations (full minus ablation).}\par\smallskip
{\footnotesize
\begin{tabular}{@{}lccc@{}}
\toprule
Ablation & $T$ & Gain & Paired 95\% CI \\
\midrule
w/o HA & 0.95 & 0.00249 & [0.00119, 0.00379] \\
w/o HA & 1.00 & 0.00205 & [0.00083, 0.00327] \\
w/o HA & 1.05 & 0.00394 & [0.00246, 0.00542] \\
w/o DT & 0.95 & 0.00193 & [0.00065, 0.00321] \\
w/o DT & 1.00 & 0.00235 & [0.00151, 0.00319] \\
w/o DT & 1.05 & 0.00321 & [0.00137, 0.00505] \\
\bottomrule
\end{tabular}\par}
\end{table}

\subsection{Ablation Study}
On SFN, we ablate the history-aware (HA) connection, replace the Transformer decision module (DT) with linear layers, and use a standard trajectory vanilla policy-gradient (VPG) update as a matched optimization control~\cite{VanillaPolicyGradient}. The VPG control retains the same policy architecture and terminal binary reward; all variants use matched initialization, update counts, sample pools, and held-out evaluation. In \cref{tab:AblationCombined}, the full architecture achieves the highest five-OD mean at every budget. Averaged across budgets, removing HA and DT reduces the mean by 0.0028 and 0.0025, respectively. GPG and VPG agree within 0.0002 at every budget, confirming the matched likelihood-ratio control.

Paired common-pool analysis confirms the contributions of both representation components. Relative to the full model, removing HA reduces the five-OD mean by 0.00205--0.00394 across budgets, while replacing DT reduces it by 0.00193--0.00321. All six 95\% confidence intervals exclude zero and remain significant after Holm correction; \cref{tab:paired_stats} reports the detailed intervals.

\subsection{Paired Statistical Analysis}
Across all six network-budget settings, GPG-HT outperforms the strongest competing baseline with statistically significant paired gains: 2.82--3.27 percentage points on SFN and 0.359--1.042 percentage points on Anaheim. Significance is based on paired 95\% confidence intervals over the ten common-pool differences, conditional on the selected checkpoints. All 36 baseline-by-setting paired tests remain significant after Holm correction. On Anaheim at $T=0.95$, GPG-HT improves over SEGAC by 0.00359, with a paired 95\% confidence interval of [0.00122, 0.00597]. The component comparisons in \cref{tab:paired_stats} likewise confirm contributions from HA and DT.

\subsection{Computational Efficiency}
Experiments run on an 11th Gen Intel Core i7-11800H CPU and an NVIDIA GeForce RTX 3070 Laptop GPU\@. The 256-dimensional GPG-HT has 7.45M parameters on SFN and 8.09M on Anaheim. Each decision-state forward pass takes 4.5--4.7 ms with peak CUDA memory below 41 MB. These measurements support millisecond-scale online routing while retaining the full history-conditioned representation.

% \subsection{Discussion}
% The results establish the mechanism of route-prefix learning in correlated networks. Realized early-edge times alter the conditional reliability of downstream alternatives; correlation supplies the predictive signal, and preserved observation-position correspondence supplies its sequential structure. Cross-attentive temporal summaries, positionwise time residuals, and sequence attention transform this structure into decision context. Paired architecture ablations confirm significant contributions from both the history-aware connection and the Transformer decision module.

% The same representation principle is relevant to graph-constrained decisions in which event identity, value, order, goal, and resource state jointly inform later actions. Within SOTA routing, GPG-HT incorporates this structure as historical context for online decisions under correlated travel times.

\section{Conclusion}
GPG-HT learns decision-relevant patterns from graph-indexed route prefixes for SOTA routing. Cross-attention between edge and time embeddings, positionwise time residuals, and ordered sequence encoding transform variable-length observations into trajectory memory. Decoder attention integrates this memory with the current decision context to predict a distribution over feasible next edges. The history-conditioned GPG objective trains the representation from terminal on-time outcomes.

Controlled experiments with correlated and independent links, together with shuffled-history and no-history controls, attribute the history gain to preserved observation-position relationships under correlated link times. Across the Sioux Falls and Anaheim benchmarks, GPG-HT achieves statistically significant paired common-pool gains over the strongest baseline in all six settings: 2.82--3.27 percentage points on SFN and 0.359--1.042 percentage points on Anaheim. Architecture ablations confirm the contributions of the history-aware connection and Transformer decision module. Together, these findings establish graph-indexed route-prefix learning as an effective representation mechanism for correlated SOTA routing.

% \clearpage
\bibliographystyle{elsarticle-num}
\bibliography{main}
\end{document}